\documentclass[letterpaper, 10 pt, conference]{ieeeconf} 

\usepackage{graphicx}
\usepackage{epstopdf}
\usepackage{subfigure}
\usepackage{times}
\usepackage{url}
\usepackage[bookmarks=true, bookmarksopen=true, bookmarksnumbered=true]{hyperref}
\hypersetup{hidelinks}
\usepackage[font=footnotesize, labelsep=colon]{caption}
\usepackage{bm}
\usepackage{amssymb}
\usepackage{amsmath}
\usepackage[ruled]{algorithm2e}
\usepackage{algorithmic,float}
\usepackage{multirow}
\usepackage{booktabs}
\usepackage{array}
\usepackage{threeparttable}
\usepackage{amsmath}
\usepackage{hyperref}
\hypersetup{
	colorlinks=true,
	linkcolor=blue,
	filecolor=blue, 
	urlcolor=blue,
	citecolor=cyan,
}

\IEEEoverridecommandlockouts                                         

\title{\LARGE \bf
PSF-LO: Parameterized Semantic Features Based Lidar Odometry
}

\author{Guibin Chen$^{*}$, Bosheng Wang$^{*}$, Xiaoliang Wang, Huanjun Deng, Bing Wang and Shuo Zhang
\thanks{All the authors are serving the Alibaba DAMO Academy Autonomous Driving Lab, Hangzhou 311121, China. $^{*}$Guibin Chen and $^{*}$Bosheng Wang are co-first authors. Corresponding author: Guibin Chen, email: {\tt\small guibin.cgb@alibaba-inc.com}.}
\thanks{This work was supported by the National Key R\&D Program of China under Grant 2018YFB1600804 and Zhejiang Program in Innovation, Entrepreneurship and Leadership Team (2018R01017).}
}

\begin{document}

\maketitle
\thispagestyle{empty}
\pagestyle{empty}

\begin{abstract}

Lidar odometry (LO) is a key technology in numerous reliable and accurate localization and mapping systems of autonomous driving. The state-of-the-art LO methods generally leverage geometric information to perform point cloud registration. Furthermore, obtaining the point cloud semantic information describing the environment more abundantly will facilitate the registration. We present a novel semantic lidar odometry method based on self-designed parameterized semantic features (PSFs) to achieve low-drift ego-motion estimation for autonomous vehicle in real time. We first use a convolutional neural network-based algorithm to obtain point-wise semantics from the input laser point cloud, and then use semantic labels to separate road, building, traffic sign and pole-like point cloud and fit them separately to obtain corresponding PSFs. A fast PSF-based matching enables us to refine geometric features (GeFs) registration, thereby reducing the impact of blurred submap surface on the accuracy of GeFs matching. Besides, we design an efficient instance-level method to accurately recognize and remove the dynamic objects while retaining static ones in the semantic point cloud, which are beneficial to further improve the accuracy of LO. We evaluate our method, namely PSF-LO, on the public dataset KITTI Odometry Benchmark and rank \#1 among semantic lidar methods with an average translational error of 0.82\% in the test dataset.

\end{abstract}

\section{\textsc{Introduction}}

Autonomous driving technology has gained major advances in recent years, where localization and mapping system is one of the essential modules. This system usually utilizes technologies based on simultaneous localization and mapping (SLAM), and the lidar/visual odometry is a fundamental module for SLAM or inertial odometry/SLAM integrated with inertial measurement unit (IMU). Therefore, it's extremely important to improve the accuracy of odometry system and reduce its cumulative drift error as much as possible. 3D lidar has advantages of high ranging accuracy and being able to work at night, so our work focuses on developing lidar odometry (LO) algorithm.

Most recent LO systems follow the frame-to-submap GeF-based matching framework, whose accuracy may suffer from excessive noisy points on submap surface \cite{deschaud2018imls} caused by the error of previously estimated poses or the quality of point cloud. In addition, dynamic objects (car, pedestrian, etc) will interfere with matching accuracy, but directly removing all objects will lose the advantage of static objects for matching \cite{chen2019suma++}. The development of deep neural networks has boosted semantic segmentation of 3D point cloud, promoting the trend of exploring semantic LO/SLAM aiming to better register a new frame to submap with the aid of semantics, which will hopefully address the above challenges.

\begin{figure}[tbp] 
  \centering 
  \includegraphics[scale=0.20]{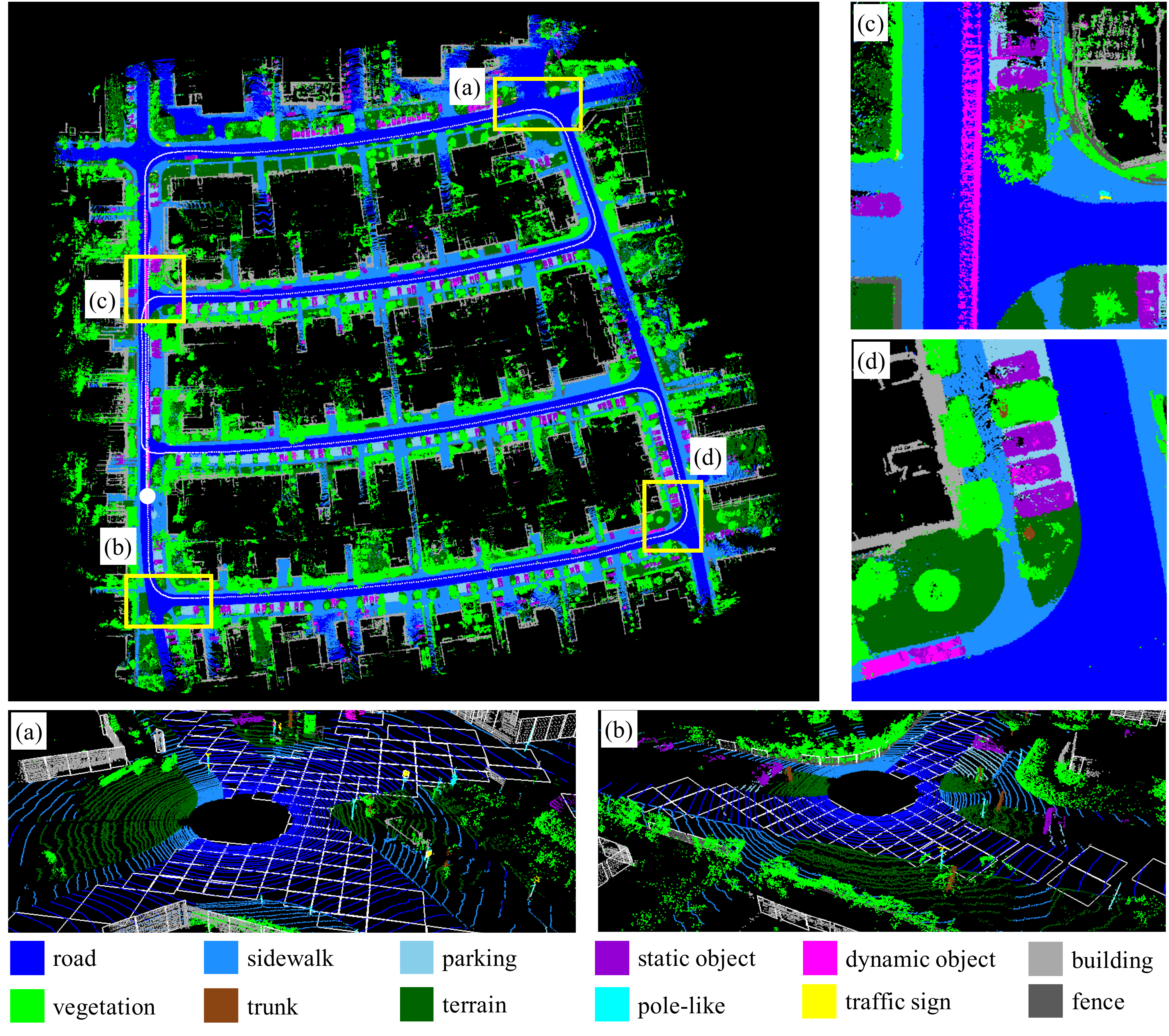} 
  \caption{A semantic map built by PSF-LO on Sequence 16 of test dataset in KITTI Odometry Benchmark. The poses are denoted as white dotted line and a big white circle indicates the starting position. (a) and (b) illustrate the PSFs extracted from single semantic point cloud rendered by their outlines. The example results of distinguishing dynamic and static objects colored in magenta and purple respectively are shown in submap (c) and (d).}
  \label{fig:example}
\end{figure}

In this paper, we present a LO method based on parameterized semantic features (PSFs) as called PSF-LO, with the view of exploring the refinement capability of PSF on GeF-based registration and the improvement of LO accuracy via removing dynamic objects while retaining static ones. Fig. \ref{fig:example} shows an example mapping result of Sequence 16 on KITTI Odometry Benchmark \cite{geiger2012we} which is based on poses calculated with PSF-LO, and the semantics are inferred by \cite{milioto2019rangenet++}. More visualization details can be seen in the attached video.

To summarize, our main contributions are as follows:
\begin{itemize}
  \item We novelly design four types of lightweight PSFs: road, building, traffic sign and pole-like, to parameterize their respective semantic point cloud abstractly and precisely.
  \item We jointly use PSFs and GeFs that compensate for shortcomings from the other to achieve a low-drift and real-time LO system aiming to improve LO accuracy by leveraging semantic information besides geometrics.
 \item We design a novel dynamic and static object classifier while considering object's temporal heading consistency, velocity magnitude and uncertainty to recognize and remove the dynamic ones during registration thereby further eliminating the LO drift.
\end{itemize}

\section{\textsc{Related Works}}

There are hundreds of literatures researching odometry/SLAM based on different types of sensors such as 2D/3D lidar, monocular/stereo/RGB-D camera, GNSS, dead reckoning sensors or mixed use of these sensors. Here we mainly concentrate on the lidar-only methods.

\textbf{Geometrics-based method.} Benefiting from centimeter-level ranging accuracy of lidar sensor, the geometric structure of point cloud is its most basic and reliable characteristic. Geometrics-based lidar odometry/SLAM can be divided into feature-based and grid-based matching methods. Concerning the former, LOAM \cite{zhang2017low} as a typical work extracts corner and surface features from the input point cloud to perform point-to-line and point-to-plane ICP registration. IMLS-SLAM \cite{deschaud2018imls} takes a special sampling strategy to select the samples on planar surfaces or far from the lidar center and then do scan-to-IMLS model matching. SuMa \cite{behley2018efficient} employs surface elements (surfels) to represent the map and perform frame-to-model ICP with point-to-plane residual. The grid-based methods such as Google's Cartographer \cite{hess2016real} and other Normal Distributions Transform-based matching algorithms \cite{schulz2020real}, \cite{ji2018cpfg}, \cite{saarinen2013normal} divide the 2D/3D space into grids at certain resolutions and do point cloud registration on those grids.

\textbf{Learning-based method.} Recently, the revolutionary development of deep learning in the image field has also promoted some explorations in lidar odometry/SLAM. \cite{cho2020unsupervised} introduces an uncertainty-aware loss with geometric confidence, thereby achieving the unsupervised deep lidar odometry. DeepPCO \cite{8967756} proposes an end-to-end lidar odometry in a supervised manner, using two parallel sub-networks to infer 3D orientation and translation, respectively. There are also some networks aiming at learning features or descriptors from the input point cloud \cite{yew20183dfeat}, \cite{lu2019deepvcp}, \cite{zhou2018voxelnet}. However, learning-based LO/SLAM methods have not yet outperformed conventional geometrics-based methods.

\textbf{Semantics-aided method.} Semantic information of point cloud can help autonomous vehicles obtain a fine-grained understanding of the surrounding environment \cite{behley2019semantickitti}. LeGO-LOAM \cite{shan2018lego} separates ground points from the point cloud used in planar features extraction and first step pose optimization. \cite{9197261} presents a hierarchical semantic mapping framework for collaborative robots or simple robots. SuMa++ \cite{chen2019suma++} exploits semantics to filter dynamic objects in a surfel level and perform a semantic ICP. In contrast to SuMa++, we concentrate on leveraging self-designed PSF to refine the GeF matching and classifying the static and dynamic state of obstacle points in an instance level.

For odometry/SLAM system, removing dynamic objects while retaining static ones can help lead better performance \cite{chen2019suma++}. In fact, distinguishing object state accurately is in the perception domain, which can be mainly divided into grid-level based and object-level based methods. The former directly detects dynamic objects on the grid map. \cite{petrovskaya2009model}, \cite{wojke2012moving} judge whether point cloud of one grid is dynamic based on the difference of virtual ray map. There are also some works \cite{tanzmeister2014grid}, \cite{nuss2018random} seeking to estimate the grid's speed to further distinguish static and dynamic states of one grid in map. \cite{wu2020motionnet} achieves the same purpose with the aid of deep networks. The latter object-level methods are typically presented in a tracking by detection architecture. \cite{darms2009obstacle}, \cite{cho2014multi}, \cite{mertz2013moving}, \cite{apollo} first extract features from objects to compose the box-model, then leverage velocity estimated by gaussian filter to distinguish dynamic and static states of objects. \cite{held2016robust} and \cite{wang2019precise} propose a state estimation system based on point level matching, overcoming the influence of box-model instability, however, the classification of static and dynamic objects is not further discussed. Our dynamic and static object classification module follows \cite{held2016robust} and integrates classifier into state estimator.

\begin{figure*}[htbp]
	\centering
	\includegraphics[scale=0.35]{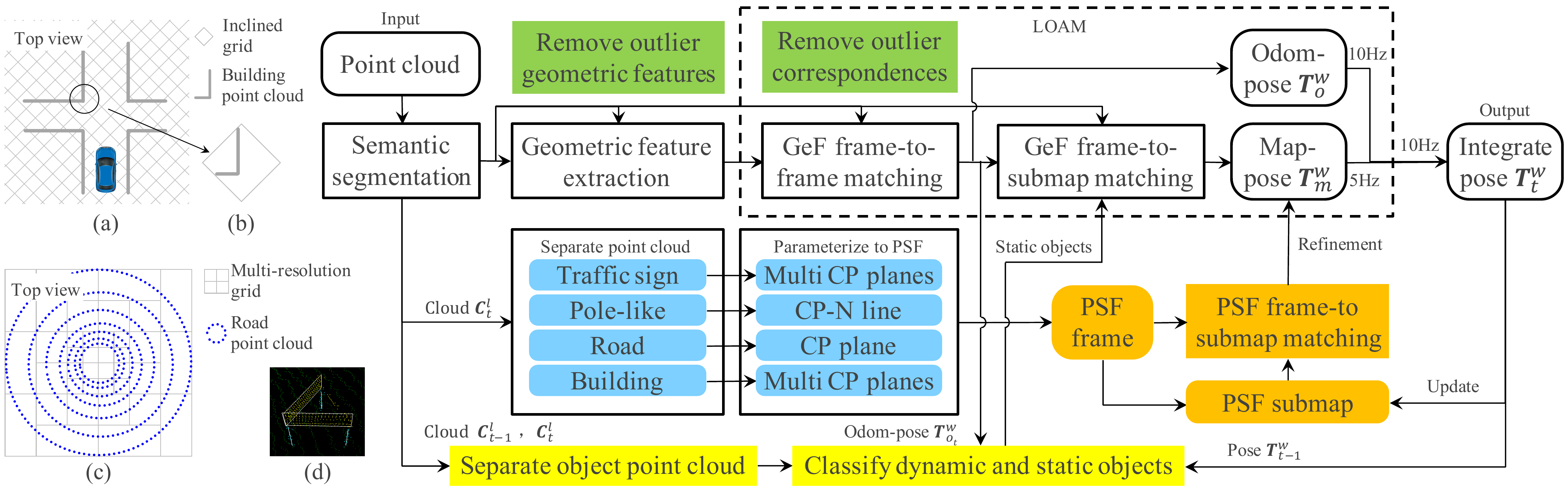} 
	\caption{Block diagram of the proposed lidar odometry method based on parameterized semantic features, namely PSF-LO.}
	\label{fig:pipeline}
\end{figure*}

\section{\textsc{Proposed Method}}

We denote the rotation and translation of a rigid transformation by a $3\times3$ matrix $\bm{R} \in SO(3)$ and a $3\times1$ translation vector $\bm{t}$ separately. $(\cdot)_{t}$ represents a variable at discrete time $t$. $(\cdot)^{w}$ and $(\cdot)^{l}$ indicate in the world and local coordinate system, and the latter is lidar coordinate system calibrated by external parameters, where the $X,Y,Z$-axis point forward, left and upward respectively. Besides, pose $\bm{T}_{t}^{w} \in SE(3)$ represents the transformation of local coordinate system at time $t$ relative to its first. $\bm{T}_{t}^{w}$ is also the ego-motion that our method aims to estimate finally, whose rotational and translational parts are denoted as $\bm{R}_{t}^{w}$ and $\bm{t}_{t}^{w}$.

\subsection{Overview}

The pipeline of proposed PSF-LO is shown in Fig. \ref{fig:pipeline}, inputting a sequence of 3D point clouds and outputting the corresponding estimated poses. We first use RangeNet++ \cite{milioto2019rangenet++} to perform semantic segmentation, and the following GeF extraction and matching mainly follow LOAM \cite{zhang2017low}. Then we extract PSFs from road, building, traffic sign and pole-like point clouds to refine the GeF matching. Besides, the dynamic and static object classification module identifies static ones who will participate in GeF matching and then be added to the GeF submap. The semantic information also helps to filter out outlier geometric features and correspondences. Let's introduce the above modules in detail.

\subsection{PSF Extraction}

The PSF extraction process is shown in Fig. \ref{fig:pipeline}, where each PSF is defined by a set of parameterized equation coefficients $\bm{c}_s \in \mathbb{R}^3 \text{ or } \mathbb{R}^6$, a weight $w_s \in \mathbb{R}$ indicating the reliability, a semantic label $l_s \in \{road, building, sign, pole\}$, a bounding box center $\bm{e}_s \in \mathbb{R}^3$ and a set of bounding box outline coefficients $\bm{o}_s$ used for rendering. The input is a semantic point cloud $\bm{C}_{t}^{l}$ and the output is a PSF-frame composed of all extracted PSFs. We first separate four types of point cloud from $\bm{C}_{t}^{l}$, and then proceed as follows,

\textbf{Road PSF:} To overcome the problem that road point cloud is dense near but sparse in the distance, we use a multi-resolution 2D grid as shown in Fig. \ref{fig:pipeline} (c) to divide the space into series of grids. Then we perform RANSAC plane fitting on road points falling in each grid and further obtain plane equation coefficients $\bm{c}_s = \left[\pi_1, \pi_2, \pi_3\right]^\mathrm{T} = d_{\pi}\bm{n}_{\pi}$ in form of closest point (CP form \cite{yang2019observability}), where $d_{\pi}$ is the scalar distance from origin to plane and $\bm{n}_{\pi}$ is the unit normal vector pointing from origin to plane. The weight $w_s$ is calculated by the ratio of inliers in total points during fitting. Next, we calculate the minimum enclosed rectangle of all inliers, taking its center as $\bm{e}_s$ and four corners as $\bm{o}_s$, and set the $l_s$ to $road$ to complete the road PSF extraction.

\textbf{Building PSF:} For large-scale building point cloud, we also need to segment it using a 2D grid, but to prevent a wall from being divided into two parts along its tangential direction, we first rotate the grid by $45^{\circ}$ as illustrated in Fig. \ref{fig:pipeline} (a). The following process is similar to road PSF, except that multiple planes may appear in one inclined grid as shown in Fig. \ref{fig:pipeline} (b). So we perform multiple planar PSFs extractions in every building grid, where the inliers are removed after once extraction and the procedure stops until the number of remaining points is less than a certain threshold.

\textbf{Traffic sign PSF:} Because the point cloud of traffic sign or pole-like below is usually distributed in isolation, we directly use Euclidean clustering to segment it. The traffic sign PSF extraction is almost the same as that of building PSF, also employing multi-plane extraction to deal with situation of multi-surface sign as shown in Fig. \ref{fig:pipeline} (d).

\textbf{Pole-like PSF:} After Euclidean clustering on pole-like point cloud, we implement 3D line RANSAC fitting for each cluster. Inspired by the CP form of plane in \cite{yang2019observability}, we design a CP-N form for 3D line representation: $\bm{c}_s = \left[p_1, p_2, p_3, n_1, n_2, n_3\right]^\mathrm{T} = \left[\bm{p}_p, \bm{n}_p\right]$, where $\bm{p}_p$ is the point closest to origin on line and $\bm{n}_p$ is the unit direction vector of line pointing to positive direction of $Z$-axis. $l_s$ is set to $pole$ and $w_s, \bm{e}_s, \bm{o}_s$ are similar to road PSF.

Fig. \ref{fig:example} (a) and (b) show the rendering effect of example PSF-frames leveraging the variable $\bm{o}_s$ of all PSFs.

\subsection{Lidar Odometry Based on PSFs}

It should be noted that the current types of PSFs are limited, so we need to combine them with more robust GeFs, thereby refining the GeF-based matching result. As shown in dotted framework of Fig. \ref{fig:pipeline}, the GeF portion of PSF-LO follows LOAM whose details can be found in \cite{zhang2017low} for the sake of brevity.

We first apply semantic information to make two simple improvements to LOAM: (1) We filter out unreliable corner GeFs extracted from the $road$ or $terrain$ point cloud with plane distribution; (2) We add a semantic consistency penalty to GeF weight $w$ in \cite{zhang2017low} :
\begin{equation}
	\label{eq:w_res}
	w = w * \frac{1.0}{1.0 + e^{-2 * N_s}}
\end{equation}
where $N_s$ is the number of target GeFs with the same semantic label as source GeF in the KNN (K-nearest neighbors) search process when establishing correspondence in GeF matching. Therefore, the weight $w$ will decrease with poor semantic consistency.

\begin{figure}[tbp] 
	\centering
	\includegraphics[scale=0.30]{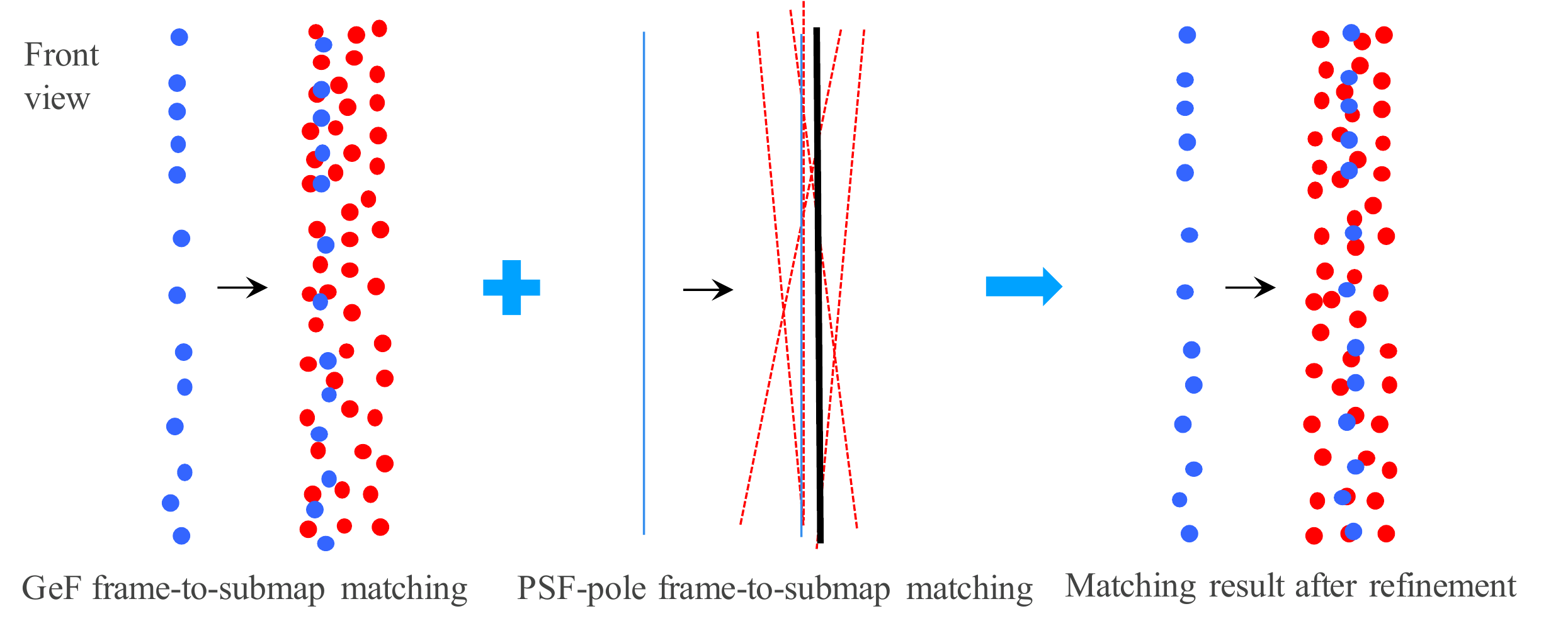} 
	\caption{Refinement of PSF to GeF registration. The blue and red points represent source and target corner features. The blue, red and black lines indicate pole-like PSF-frame, PSF-submap and weighted average PSF.}
	\label{fig:refine}
\end{figure}

Then Fig. \ref{fig:refine} shows the refinement principle of pole-like PSF on GeF-based registration, and the refinement principles of other types of PSFs are similar. The PSF-submap is composed by sequences of recent PSF-frames which are within 100m to current frame and have been transformed into world coordinate system according to previous estimated poses. And the transformation of road, building and traffic sign PSFs which belong to planar PSFs with a given pose
\begin{math}
\bm{T}^{w} =
\left[
\begin{smallmatrix}
\bm{R}^{w} & \bm{t}^{w}\\
0 & 1\\
\end{smallmatrix}
\right]
\end{math} can be computed as
\begin{equation}
\label{eq:psf_trans}
	\begin{cases}
		w_s^w = w_s^l \text{, } l_s^w = l_s^l \text{, } \bm{e}_s^w = \bm{R}^w \bm{e}_s^l + \bm{t}^w \\
		\bm{o}_{s}^w = \bm{R}^w \bm{o}_{s}^l + \bm{t}^w \text{, } \bm{c}_s^w = d_{\pi}^w \bm{n}_{\pi}^w
	\end{cases}
\end{equation}
where $d_{\pi}^w$ and $\bm{n}_{\pi}^w$ can be derived by $\bm{c}_s^l$ \cite{geneva2018lips},
\begin{equation}
    \label{eq:trans_n}
	\begin{bmatrix}
	\bm{n}_{\pi}^w \\
	d_{\pi}^w
	\end{bmatrix}
	=
	\begin{bmatrix}
	\bm{R}^w & 0 \\
	((\bm{R}^w)^{-1}\bm{t}^w))^\mathrm{T} & 1
	\end{bmatrix}
	\begin{bmatrix}
	{\bm{c}_s^l} / {||\bm{c}_s^l||_{2}} \\
	||\bm{c}_s^l||_{2}
	\end{bmatrix} \\
\end{equation}
and the pole-like PSF transformation is similar to planar PSF, only $\bm{c}_s^w = \left[\bm{p}_p^w, \bm{n}_p^w\right]$ is different, where
$\bm{p}_p^w = \bm{R}^w \bm{p}_p^l + \bm{t}^w \text{, }
  \bm{n}_p^w = \bm{R}^w \bm{n}_p^l$.

Let $\bm{T}_m^w$ be the map-pose which is initialized by the odom-pose $\bm{T}_o^w$, where $\bm{T}_o^w$ and $\bm{T}_m^w$ are estimated by the frame-to-frame and frame-to-submap GeF matching respectively as shown in Fig. \ref{fig:pipeline}. And for clarity, we individually denote the PSF, PSF-frame and PSF-submap as $\mathcal{S}$, $\mathcal{F}_s$ and $\mathcal{Q}_s$. Besides, we use $\mathcal{S} \langle \cdot \rangle$ to indicate a certain variable of $\mathcal{S}$.

In order to find the corresponding PSFs for each $\mathcal{S}^l$ in $\mathcal{F}_s^l$ from $\mathcal{Q}_s^w$, we store the center $\bm{e}_s^w$ of all PSFs in $\mathcal{Q}_s^w$ into a Kd-tree \cite{van2000computational} and transform $\mathcal{F}_s^l$ into $\mathcal{F}_s^w$ using $\bm{T}_m^w$ which is denoted as $\mathcal{F}_s^w = \bm{T}_m^w \boxtimes \mathcal{F}_s^l$, where $\boxtimes$ represents the aforementioned PSF transformation. Then for each $\mathcal{S}^w$ in $\mathcal{F}_s^w$, we search for a set of same type PSFs $\mathcal{S}_\mathcal{Q}^w = \{\mathcal{S}_i^w \mid i = 1, ..., n\}$ in $\mathcal{Q}_s^w$ whose $\bm{e}_s^w$ is close to $\bm{e}_s^w$ of $\mathcal{S}^w$ within a certain range. And the weighted average PSF $\bar{\mathcal{S}}^w$ of $\mathcal{S}_\mathcal{Q}^w$ can be computed as
\begin{equation}
\label{eq:psf_avg}
	\bar{\mathcal{S}^w} \langle \bm{c}_s^w \rangle = \frac{\sum_{i=1}^{n} \mathcal{S}_i^w \langle \bm{c}_s^w \rangle * \mathcal{S}_i^w \langle w_s^w \rangle} {\sum_{i=1}^{n} \mathcal{S}_i^w \langle w_s^w \rangle}
\end{equation}
With $\bar{\mathcal{S}}^w$ calculated, we can further derive the error function between it and $\mathcal{S}^w$. But due to the singularity of planar CP form \cite{geneva2018lips} and the increase of \textit{error state} with increase of distance scale \cite{yang2019observability}, we need to calculate error in local coordinate system. As for planar PSF, the plane-to-plane error can finally be given by:
\begin{equation}
	E_{\pi} = \mathcal{S}^l \langle w_s^l \rangle * ||\mathcal{S}^l \langle \bm{c}_s^l \rangle - (\bm{T}_m^w)^{-1} \boxtimes \bar{\mathcal{S}}^w \langle \bm{c}_s^w \rangle||_2
\end{equation}

With respect to pole-like PSF, we first introduce the CP form $\bm{c}_{s'} = d_p\bar{\bm{q}}_p$ of 3D line representation in \cite{yang2019observability}, where $d_p$ is scalar distance from origin to line and $\bar{\bm{q}}_p$ is a special unit quaternion. The transformation from our CP-N form $\bm{c}_s = \left[\bm{p}_p, \bm{n}_p\right]$ to the CP form $\bm{c}_{s'} = d_p\bar{\bm{q}}_p$ is as follows
\begin{equation}
\begin{cases}
	\bm{p}_0 = \bm{n}_p + \bm{p}_p \text{, } \bm{p}_1= -\bm{n}_p + \bm{p}_p \\
	\bm{n}_0 = \bm{p}_0 \times \bm{p}_1 \text{, } \bm{n}_1 = \bm{n}_0 \times \bm{n}_p \\
	\bm{R}(\bar{\bm{q}}_p) = \left[\frac{\bm{n}_0}{||\bm{n}_0||_2}, \bm{n}_p, \frac{\bm{n}_1}{||\bm{n}_1||_2} \right] \text{, } d_p = \frac{||\bm{p}_0 \times \bm{p}_1||_2}{||\bm{p}_0 - \bm{p}_1||_2}
\end{cases}
\end{equation}
We denote the above transformation as $\bm{c}_{s'} = \lfloor \bm{c}_{s} \rfloor$, so the line-to-line error can be defined as
\begin{equation}
	E_p = \mathcal{S}^l \langle w_s^l \rangle * || \lfloor \mathcal{S}^l \langle \bm{c}_s^l \rangle \rfloor - \lfloor (\bm{T}_m^w)^{-1} \boxtimes \bar{\mathcal{S}}^w \langle \bm{c}_s^w \rangle \rfloor||_2
\end{equation}

Besides $E_{\pi}$ and $E_p$, the point-to-line error $E_{gl}$ and point-to-plane error $E_{g\pi}$ can be obtained when performing frame-to-submap GeF matching as described in detail in \cite{zhang2017low}. Stacking $E_{\pi}$, $E_p$, $E_{gl}$ and $E_{g\pi}$ to $E$, a nonlinear function $f(\bm{T}_m^w) = E$ can be derived, thereby map-pose $\bm{T}_m^w$ will be estimated by solving the nonlinear least square problem $\arg\min \sum||E||^2$. After necessary number of iterations, we can get the final refined $\bm{T}_m^w$, and integrate it with odom-pose $\bm{T}_o^w$ to output the real-time ego-motion estimation $\bm{T}_t^w$.

\subsection{Dynamic and Static Object Classification}

\begin{figure}[tbp]
	\centering
	\includegraphics[width=7.6cm,height=3.0cm]{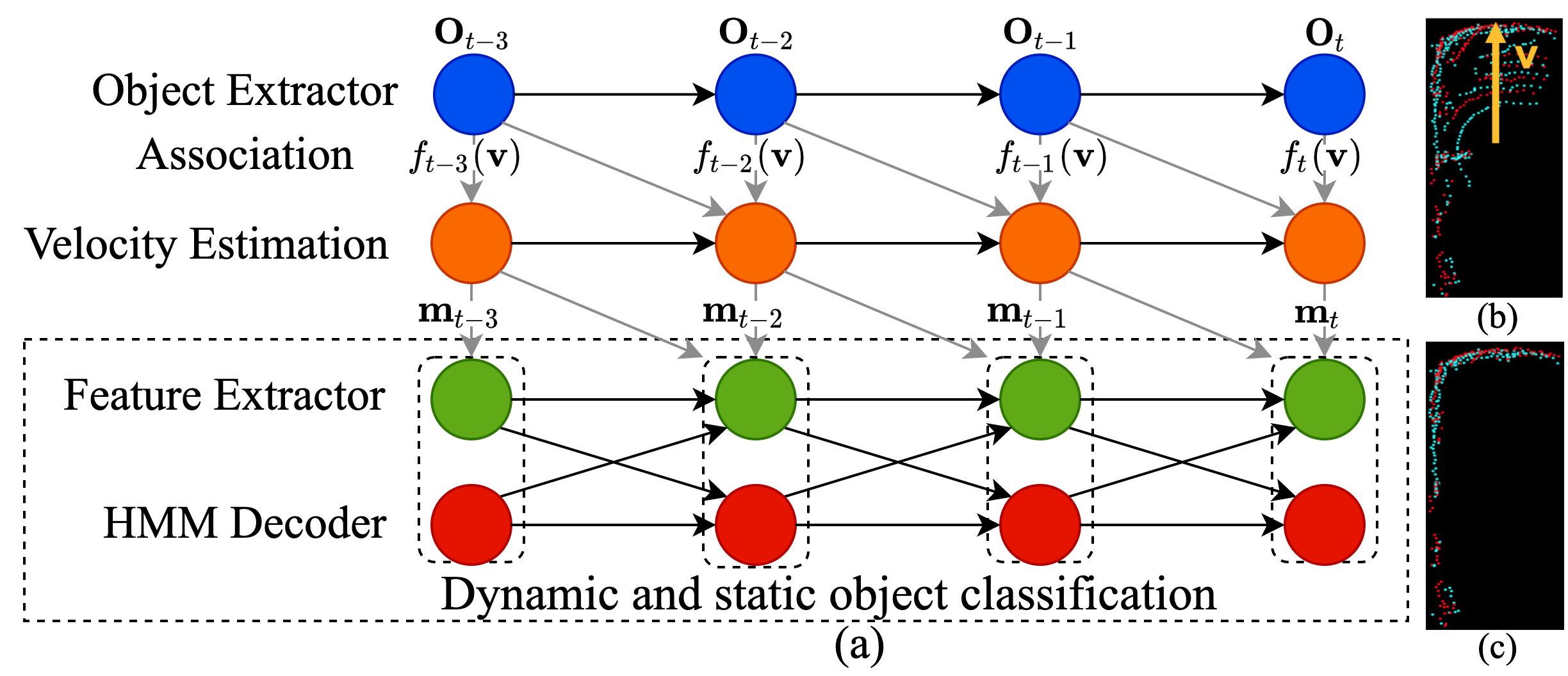} 
	\caption{(a) shows framework of our classification method, where the red and green ball represent static and dynamic states respectively. (b) and (c) show the effect of our sampling strategy. The arrow in (b) indicates a wrong velocity estimation which should be 0.}
	\label{fig:sd_fw}
\end{figure}

The proposed classification method is based on tracking-by-detection framework as shown in Fig. \ref{fig:sd_fw} (a). Our method consists of four parts: object extractor, association, velocity estimation, dynamic and static object classification. For object extractor, we use the method of \cite{bogoslavskyi2016fast} to cluster the separated object points $\mathbf{O}_t$. We utilize a greedy algorithm for data association based on the distance of object point cloud center. The current predictive pose $\bm{T}_{o_t}^w$ and the last integrated pose $\bm{T}_{t-1}^w$ are used to transform the corresponding objects to world coordinate system as shown in Fig. \ref{fig:pipeline}.

\textbf{Velocity estimation}: We use \cite{held2016robust} for velocity estimation, but the original algorithm is easily affected by the upper surface of the objects. As shown in Fig. \ref{fig:sd_fw} (b), the red and blue points are from a static object in current and last frame, then directly applying the original algorithm in \cite{held2016robust} results in a wrong velocity estimation. Consequently, a point sampling algorithm based on vertical planes is proposed to handle this issue. The point cloud of object is projected into the 2D polar coordinates according to the angular and range resolution, $\Delta\theta$, $\Delta d$. We add a tag describing the vertical state for each $bin(i,j)$. The points projected into $bin(i,j)$ are noted as $\mathbf{p}_{i,j}$. Let  $\mathbf{P}_{i,j} = \{\mathbf{p}_{m,n}, |m-i|\le r, |n-j|\le r\}$ be the set of all points projected around $bin(i,j)$, we count the max consecutive rings number $R(i,j)$ in $\mathbf{P}_{i,j}$. If $R(i, j)$ is greater than a threshold $n_v$, the bin is tagged as vertical.
 It means that we prefer the bin continuously hit by different laser rings. Then, we utilize the points with vertical tags as shown in Fig. \ref{fig:sd_fw} (c) to do velocity estimation. Finally, the probability density distribution of the velocity $f_t(\mathbf{v})$ is achieved.

 \textbf{Feature extractor}: $f_t(\mathbf{v})$ is needed to be transferred to a measurement vector, which describes the static/dynamic state: $\mathbf{m}_t = (p_{s_t}, p_{d_t}), p_{s_t} + p_{d_t} = 1.0$. Usually, velocity magnitude and heading consistency are utilized to distinguish object's state. However, only relying on the former results in missing detection of slow-moving object, on the other hand only relying on the latter introduces false positive dynamic object, especially when the point cloud matching fails. Therefore, we propose to utilize both of them to do dynamics estimation when point cloud matching is unstable. Velocity uncertainty modeling is proposed to measure the quality of point cloud matching, which is defined as $p_{u_t}$, 0 for good and 1 for bad. The score $p_{s_t}(h_\sigma)$ is calculated based on heading consistency. And the score $p_{s_t}(\mathbf{v})$ is achieved based on velocity magnitude. Then, we utilize the rules of Bayes Filter to generate the final static score as follows:
 \begin{equation}
 \label{eq:dyn2}
 \texttt{odds}(p_{s_t}) = \texttt{odds}\left(p_{s_t}(h_{\sigma})\right)+ \\
 p_{u_t} \cdot \texttt{odds}\left(p_{s_t}(\mathbf{v})\right), 
 \end{equation}
  where $\texttt{odds}(p)=\log(p(1-p)^{-1})$ and $p_{u_t}$ is defined in Eq. \ref{eq:dyn1}. $Cov(\cdot)$ is the covariance matrix fitted from $f_t(\mathbf{v})$ as \cite{held2016robust}, $Sig(x) = 1 / (1 + e^{a(b - x)})$ is utilized to normalize $\Vert Cov \Vert$. The $p_{s_t}(\mathbf{v})$ in Eq. \ref{eq:dyn2} is defined as ${Sig}_v\left({f_t(\mathbf{0})} \big/{\max_\mathbf{v} f_t(\mathbf{v})}\right)$.
 \begin{equation}
 \begin{gathered}
 \label{eq:dyn1}
 p_{u_t} = {Sig}_u\left(\left\lVert Cov(f_t(\mathbf{v}))\right\rVert \right), \lVert \mathbf{A} \rVert = \sup_{\Vert \mathbf{u} \Vert = 1} \lVert \mathbf{Au} \rVert
 \end{gathered}
 \end{equation}

 Generally speaking, the heading consistency is usually calculated based on the angle difference between the adjacent frames \cite{apollo}. However, we propose to utilize circular statistics \cite{fisher1995statistical}, \cite{berens2009circstat} to generate the distribution of heading $\theta_i$ with variance $h_\sigma = \large\sqrt{-\ln\left(\bar{C}^2 + \bar{S}^2\right)}$
, where $\bar{C}$, $\bar{S}$ are the means of $\cos\theta_i$ and $\sin\theta_i$ respectively.
To decrease the impact of noises, only the top $k$ heading directions at time $t-1$ and $t$ are selected to generate the above heading distribution. Then, the Gaussian distribution $N_d(\cdot)$ and $N_s(\cdot)$ are achieved based on the dynamic and static object's variance. The final probability of static is calculated as follows:
\begin{equation}
p_{s_t}(h_{\sigma}) = {N_{s}(h_{\sigma})}\big/\left( {N_{d}(h_{\sigma}) + N_{s}(h_{\sigma})}\right).
\end{equation}
 
 \textbf{HMM Decoder}: 
 The HMM (Hidden Markov Models) is utilized to handle the classification problem. During the LO task, the objective function is defined as: $x_t = \max_{x_{1:t}} p(x_{1:t}\lvert\mathbf{m}_{1:t})$, where $x_t$ represents if the object is dynamic or static. Viterbi algorithm \cite{forney1973viterbi} is utilized to solve the above optimization problem efficiently.

Finally, the point cloud of objects identified as static are treated as surface geometric features to participate in GeF frame-to-submap matching of PSF-LO as shown in Fig. \ref{fig:pipeline}, which will further improve the matching accuracy.

\section{\textsc{Experiments}}

We evaluate our dynamic and static objects classification method using the 3D object point cloud with point-wise ground truth semantic label on training dataset of SemanticKITTI \cite{behley2019semantickitti}, which annotates moving and non-moving objects with distinct labels. \cite{behley2019semantickitti} is actually based on KITTI Odometry Benchmark \cite{geiger2012we} which provides 22 undistorted point cloud sequences covering different scenes. The point clouds are logged from Velodyne HDL-64E mounted on a passenger vehicle. The first 11 sequences are given with RTK/IMU-based ground truth poses for algorithm training and the remaining ones are used for testing via website\footnote{\url{http://www.cvlibs.net/datasets/kitti/eval_odometry.php}\label{url:kitti}}.

Therefore we then test the performance of proposed PSF-LO on both training and test dataset of KITTI Odometry Benchmark with evaluation metric defined in \textsuperscript{\ref{url:kitti}}. The hardware condition is a GeForce RTX 2080Ti GPU and an Intel i7-8700 CPU with 4 cores @ 3.2GHz with 32GB RAM.

\subsection{Dynamic and Static Object Classification Experiment}
\begin{table}[tbp]
  \centering
  \caption{Object classification results on SemanticKITTI training dataset.}
    \begin{tabular}{ccccccc}
    \toprule
    \multicolumn{1}{c}{\multirow{2}[4]{*}{Method}} & \multicolumn{3}{c}{Static Objects} & \multicolumn{3}{c}{Dynamic Objects} \\
\cmidrule{2-7}         &
\multicolumn{1}{p{2.7em}}{Precision\newline{}(\%)} &
\multicolumn{1}{p{2em}}{Recall\newline{}(\%)} &
\multicolumn{1}{p{2em}}{IoU\newline{}(\%)} &
\multicolumn{1}{p{2.7em}}{Precision\newline{}(\%)} &
\multicolumn{1}{p{2em}}{Recall\newline{}(\%)} &
\multicolumn{1}{p{2em}}{IoU\newline{}(\%)} \\
    \midrule
    Ours base & 99.27 & \textbf{85.37} & \textbf{84.84} & \textbf{23.53} & 87.81 & \textbf{22.78} \\
    \midrule
    Ours no-s & \textbf{99.32} & 82.98 & 82.51 & 21.13 & \textbf{89.04} & 20.60 \\
    \midrule
    Ours no-u & 99.28 & 82.62 & 82.13 & 20.66 & 88.35 & 20.11 \\
    \midrule
    \midrule
    SuMa++ \cite{chen2019suma++} & 96.70 & 20.14 & 20.01 & 5.26 & 86.58 & 5.22 \\
    \bottomrule
    \end{tabular}%
  \label{tab:dyn_static}%
\end{table}%

\begin{figure}[bp]
	\centering
	\includegraphics[scale=0.4]{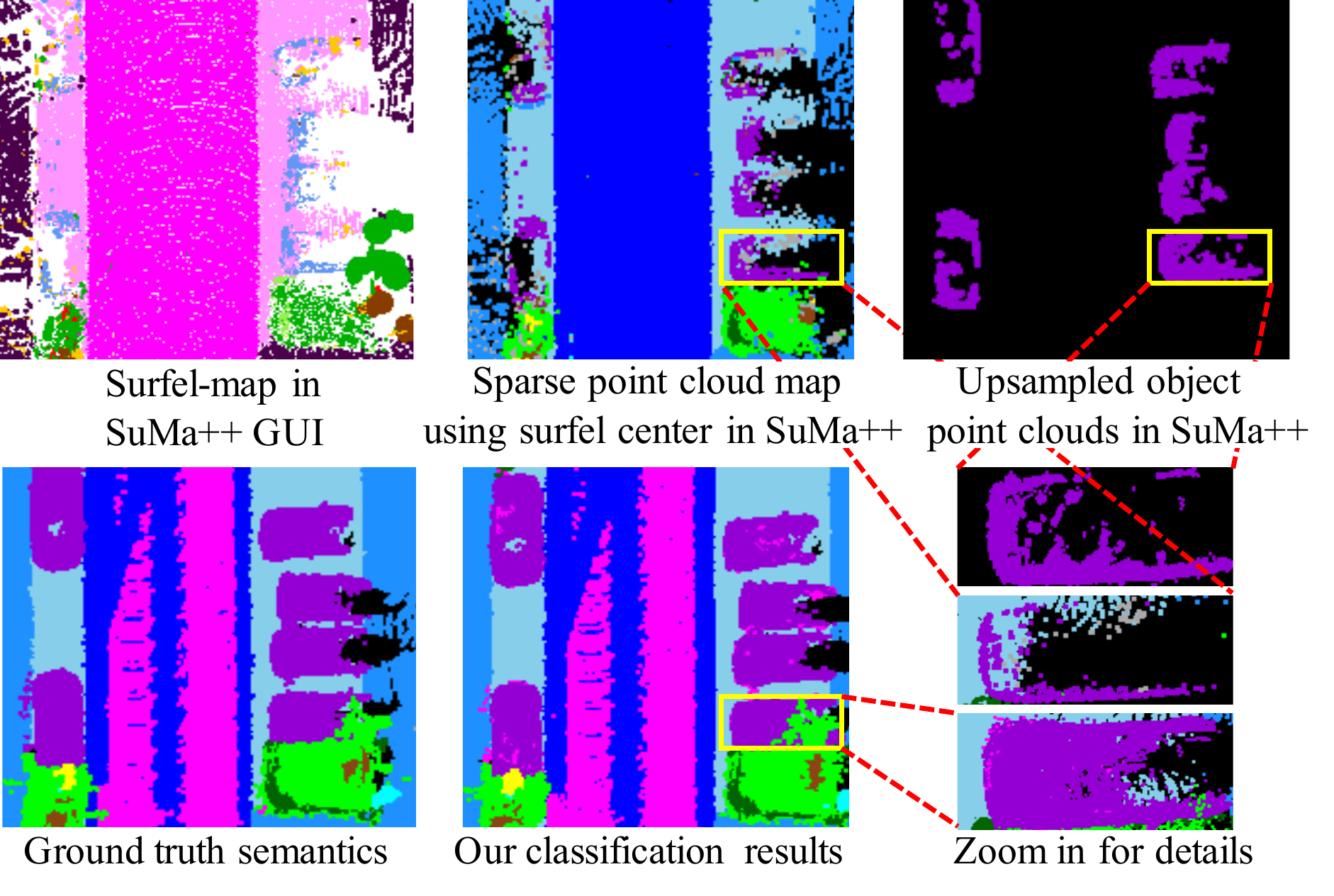} 
	\caption{Visualization of object classification results on Sequence 07 of SemanticKITTI. Note that SuMa++ GUI is colored differently than others.}
	\label{fig:dyn_cmp}
\end{figure}

\begin{table*}[tbp]
	\scriptsize
	\centering
	\caption{Translational error comparison of several different advanced lidar odometry/SLAM methods on KITTI Odometry Benchmark.}
	\begin{tabular}{cccccccccccccc}
		\toprule
		\multirow{2}[4]{*}{Method} & \multicolumn{12}{c}{Sequences on training dataset}                                 & Test dataset \\
		\cmidrule{2-14}         & 00   & 01   & 02   & 03   & 04   & 05   & 06   & 07   & 08   & 09   & 10   & \multirow{1}[2]{*}{Average} & \multirow{1}[2]{*}{Average} \\
		& Urban & Highway & Urban & Country & Country & Country & Urban & Urban & Urban & Urban & Country &      &  \\
		\midrule
		LOAM \cite{zhang2017low} & 0.78  & 1.43  & 0.92  & 0.86  & 0.71  & 0.57  & 0.65  & 0.63  & 1.12  & 0.77  & 0.79  & 0.87  & 0.88  \\
		\midrule
		IMLS-SLAM \cite{deschaud2018imls} & 0.50  & 0.82  & 0.53  & 0.68  & 0.33  & 0.32  & 0.33  & 0.33  & 0.80  & 0.55  & 0.53  & 0.55  & 0.69  \\
		\midrule
		SuMa \cite{behley2018efficient} & 0.68  & 1.70  & 1.20  & 0.74  & 0.44  & 0.43  & 0.54  & 0.74  & 1.20  & 0.62  & 0.72  & 0.83  & 1.39  \\
		\midrule
		$\blacklozenge$ SuMa++ \cite{chen2019suma++} &  \textbf{0.64}  & 1.60  & 1.00  &  \textbf{0.67}  &  \textbf{0.37}  &  \textbf{0.40}  &  \textbf{0.46}  &  \textbf{0.34}  & 1.10  &  \textbf{0.47}  & 0.66  &  \textbf{0.70}  & 1.06  \\
		\midrule
		\midrule
		{$\blacklozenge$ PSF-LO no-o} & 0.65  & \textbf{1.32}  & 0.92  & 0.76  & 0.65  & 0.45  & 0.48  & 0.45  & 0.95  & 0.55  & 0.56  & 0.76  & \textbackslash{} \\
		\midrule
		$\blacklozenge$ PSF-LO $\left[\text{Ours}\right]$& \textbf{0.64}  & \textbf{1.32}  & \textbf{0.87}  & 0.75  & 0.66  & 0.45  & 0.47  & 0.46  & \textbf{0.94}  & 0.56  & \textbf{0.54}  & 0.74  & \textbf{0.82}  \\
		\bottomrule
	\end{tabular}
	\label{tab:psf-lo}
	\begin{tablenotes}
		\item The symbol $\blacklozenge$ indicates semantic lidar methods. And the errors are measured in percent (\%), demonstrating an average drift in one hundred meters in three directions along the $X, Y, Z$-axis. The bold numbers represent top performance for semantic lidar methods.
	\end{tablenotes}
\end{table*}

We first explore impact of different core parts of our classification method. The quantitative experimental results are shown in Table \ref{tab:dyn_static}, where we use the provided ground truth semantics to calculate the point-wise average classification precision, recall rate and IoU of the dynamic and static objects. Furthermore, considering the semantic segmentation errors, we only calculate the above metrics in the true positive segmentation results of objects. Our ``base'' version represents the proposed method with all core parts. The parameters of vertical plane points sampling are set as $(\Delta\theta, \Delta d, r, n_v)\rightarrow(0.5^\circ, 0.025m, 1, 2)$. The core parameter $k$ is set to 5. The transition matrix in HMM $M = \{0.99, 0.01; 0.99, 0.01\}$. ${Sig}_u \rightarrow (a=2.0, b=1.0)$, ${Sig}_v\rightarrow(a=100, b=0.05)$, $N_{s}\rightarrow(\mu=2.06, \sigma=0.35)$, $N_{d} \rightarrow (\mu=0.05, \sigma=0.53)$. Furthermore, we set short track with length less than three as dynamic. ``no-s'' version is without vertical planes points sampling, ``no-u'' version represents that only heading consistency is utilized.

Table \ref{tab:dyn_static} shows that vertical plane points sampling brings obvious improvements for dynamic object recognition. Only utilizing heading consistency results in large amount of false alarms for dynamic objects.

We then compare the classification results with SuMa++ \cite{chen2019suma++} as shown in the bottom of Table \ref{tab:dyn_static}. Through its open source code, we can only obtain the sparse surfel-based map where the dynamic objects have been removed, so we think that the preserved and removed object point clouds are judged to be static and dynamic respectively by it. And to be fair, we leverage the original dense point cloud to upsample the sparse map by adding original object points falling within the radius of corresponding preserved surfel to the sparse map. We can see that our classification method outperforms SuMa++ under all metrics. Fig. \ref{fig:dyn_cmp} also demonstrates the comparison results qualitatively. The simple surfel-level binary Bayes Filter used in SuMa++ causes that a large number of static object points (surfels) are mistakenly removed.

Note that our average dynamic precision rate is only 23.53\% as shown in Table \ref{tab:dyn_static}, mainly due to the serious unbalanced ratio (about 19.5:1) of static and dynamic object points in the training dataset of SemanticKITTI.

\subsection{PSF-LO on KITTI Odometry Benchmark}

The second experiment is to verify the accuracy of PSF-LO on KITTI Odometry Benchmark, which finally achieves average translational error of 0.74\%/0.82\% and rotational error of 0.0027/0.0032deg/m on the training/test dataset. PSF-LO is ranked \#1 among semantic lidar methods as shown on the benchmark website\textsuperscript{\ref{url:kitti}}.

In Table \ref{tab:psf-lo}, we compare our PSF-LO with several advanced lidar methods with the main metric of translational error on KITTI training and test dataset. We can see that PSF-LO achieves lower errors than the original LOAM based on only GeF matching, which proves the refinement effect of exploiting semantics. Note that the results of LOAM are obtained from \cite{zhang2017low} which were later improved and ranked top among all lidar methods at present, but the subsequent improvement measures have not been made public. Our method also outperforms SuMa \cite{behley2018efficient} which only constructs point-to-plane constraints in registration. Both of SuMa++ \cite{chen2019suma++} and our PSF-LO belong to semantic lidar methods, but our multiple types of constraints and more accurate object state classification make us surpass it on the test dataset and approach it on the training dataset, which indicates that our method has better generalization ability on unknown scenes. In comparison with non-real-time IMLS-SLAM \cite{deschaud2018imls} which takes at 1.25 s per frame, we leverage the refinement capacity of lightweight PSFs to solve the problem of excessive noises existing on submap surface more efficiently and achieve LO accuracy close to it. The high and low frequency dual-thread structure can enable PSF-LO to realize real-time ego-motion estimation. The specific processing time of PSF-LO will be introduced in Section \ref{sec:discussion}.

Compared with PSF-LO, the ``no-o" version in Table \ref{tab:psf-lo} removes all objects without state classification and performs worse on most sequences, which proves that retaining sufficient static objects is beneficial to LO.

\subsection{Discussion}
\label{sec:discussion}
\begin{figure}[bp]
	\centering
	\includegraphics[scale=0.25]{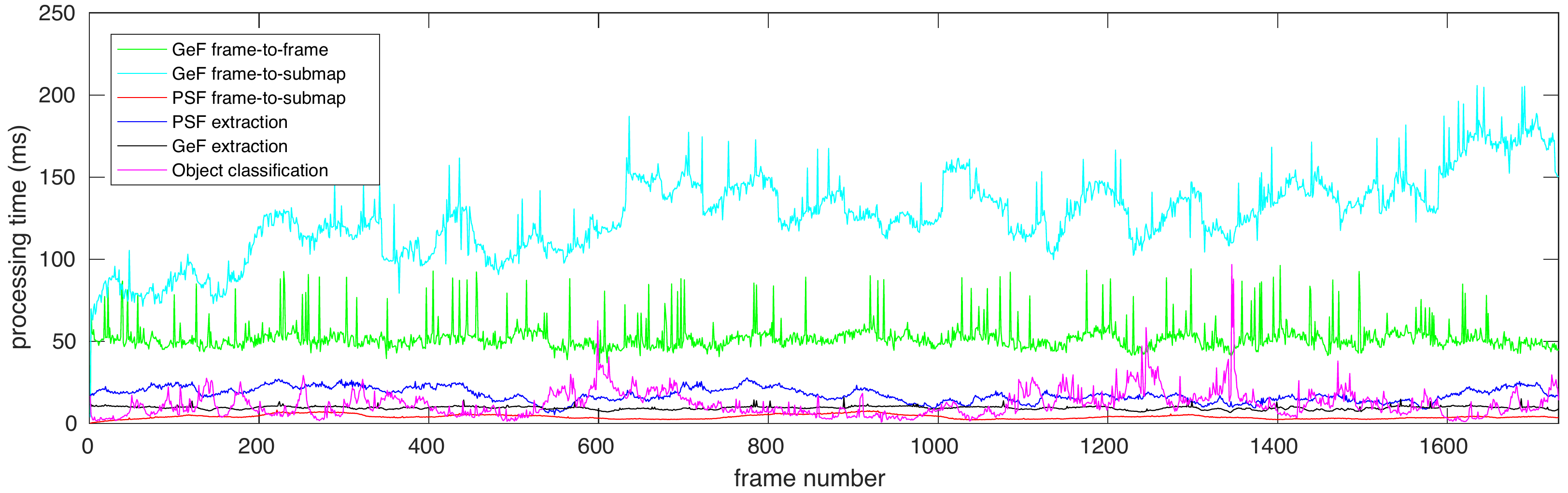} 
	\caption{Processing time of each module in PSF-LO on Sequence 16 of KITTI Odometry Benchmark. The final semantic map is as shown in Fig. \ref{fig:example}.}
	\label{fig:time}
\end{figure}
The processing time of each module in PSF-LO running on Sequence 16 of KITTI test dataset is summarized in Fig. \ref{fig:time}. We use four threads to process in parallel: one for semantic segmentation by RangeNet++ whose runtime is omitted and details can be found in \cite{milioto2019rangenet++}; one for extraction of GeFs and PSFs as well as classification of dynamic and static object, which take an average of 10, 20, 15 ms per frame respectively; one for GeF frame-to-frame matching that needs on average 50 ms; and the last one for frame-to-submap matching of GeF and PSF, which takes the most time due to the downsampling of GeF submap and construction of its Kd-tree, but our PSF matching only accounts for about 5 ms in this thread. Fast hash map \cite{hash-map} and GPU operation can be used to improve the efficiency, but currently limited by time, we reduced the publication frequency of KITTI point cloud to 5Hz to obtain better results shown in Table \ref{tab:psf-lo}.

\section{\textsc{Conclusion}}

In this paper, we propose a novel semantic lidar odometry method, namely PSF-LO, to estimate the ego-motion of vehicle, which aims to use semantic information of point cloud to improve the LO accuracy. Our main contributions are designing four types of PSFs used to refine the GeF registration and proposing a fast dynamic and static object classification algorithm to accurately recognize and retain static objects in favor of LO. We test the designed dynamic and static object classification method on SemanticKITTI Dataset so as to prove the advantage of better classification accuracy. And the experimental results on KITTI Odometry Benchmark demonstrates low drift of proposed PSF-LO in various scenarios with an average translational error of 0.74\%/0.82\% on the training/test dataset. In the future, we plan to explore more types of PSF, such as extracting PSF that can be parameterized into curves from the edge of road.


\bibliographystyle{ieeetr}
\bibliography{references} 

\begin{thebibliography}{10}

\bibitem{deschaud2018imls}
J.-E. Deschaud, ``Imls-slam: scan-to-model matching based on 3d data,'' in {\em
  2018 IEEE International Conference on Robotics and Automation (ICRA)},
  pp.~2480--2485, 2018.

\bibitem{chen2019suma++}
X.~Chen, A.~Milioto, E.~Palazzolo, P.~Gigu{\`e}re, J.~Behley, and C.~Stachniss,
  ``Suma++: Efficient lidar-based semantic slam,'' in {\em 2019 IEEE/RSJ
  International Conference on Intelligent Robots and Systems (IROS)},
  pp.~4530--4537, 2019.

\bibitem{geiger2012we}
A.~Geiger, P.~Lenz, and R.~Urtasun, ``Are we ready for autonomous driving? the
  kitti vision benchmark suite,'' in {\em 2012 IEEE Conference on Computer
  Vision and Pattern Recognition}, pp.~3354--3361, 2012.

\bibitem{milioto2019rangenet++}
A.~Milioto, I.~Vizzo, J.~Behley, and C.~Stachniss, ``Rangenet++: Fast and
  accurate lidar semantic segmentation,'' in {\em 2019 IEEE/RSJ International
  Conference on Intelligent Robots and Systems (IROS)}, pp.~4213--4220, 2019.

\bibitem{zhang2017low}
J.~Zhang and S.~Singh, ``Low-drift and real-time lidar odometry and mapping,''
  {\em Autonomous Robots}, vol.~41, no.~2, pp.~401--416, 2017.

\bibitem{behley2018efficient}
J.~Behley and C.~Stachniss, ``Efficient surfel-based slam using 3d laser range
  data in urban environments.,'' in {\em Robotics: Science and Systems}, 2018.

\bibitem{hess2016real}
W.~Hess, D.~Kohler, H.~Rapp, and D.~Andor, ``Real-time loop closure in 2d lidar
  slam,'' in {\em 2016 IEEE International Conference on Robotics and Automation
  (ICRA)}, pp.~1271--1278, 2016.

\bibitem{schulz2020real}
C.~{Schulz} and A.~{Zell}, ``Real-time graph-based slam with occupancy normal
  distributions transforms,'' in {\em 2020 IEEE International Conference on
  Robotics and Automation (ICRA)}, pp.~3106--3111, 2020.

\bibitem{ji2018cpfg}
K.~Ji, H.~Chen, H.~Di, J.~Gong, G.~Xiong, J.~Qi, and T.~Yi, ``Cpfg-slam: a
  robust simultaneous localization and mapping based on lidar in off-road
  environment,'' in {\em 2018 IEEE Intelligent Vehicles Symposium (IV)},
  pp.~650--655, 2018.

\bibitem{saarinen2013normal}
J.~Saarinen, H.~Andreasson, T.~Stoyanov, J.~Ala-Luhtala, and A.~J. Lilienthal,
  ``Normal distributions transform occupancy maps: Application to large-scale
  online 3d mapping,'' in {\em 2013 IEEE International Conference on Robotics
  and Automation}, pp.~2233--2238, 2013.

\bibitem{cho2020unsupervised}
Y.~{Cho}, G.~{Kim}, and A.~{Kim}, ``Unsupervised geometry-aware deep lidar
  odometry,'' in {\em 2020 IEEE International Conference on Robotics and
  Automation (ICRA)}, pp.~2145--2152, 2020.

\bibitem{8967756}
W.~{Wang}, M.~R.~U. {Saputra}, P.~{Zhao}, P.~{Gusmao}, B.~{Yang}, C.~{Chen},
  A.~{Markham}, and N.~{Trigoni}, ``Deeppco: End-to-end point cloud odometry
  through deep parallel neural network,'' in {\em 2019 IEEE/RSJ International
  Conference on Intelligent Robots and Systems (IROS)}, pp.~3248--3254, 2019.

\bibitem{yew20183dfeat}
Z.~J. Yew and G.~H. Lee, ``3dfeat-net: Weakly supervised local 3d features for
  point cloud registration,'' in {\em European Conference on Computer Vision},
  pp.~630--646, Springer, 2018.

\bibitem{lu2019deepvcp}
W.~Lu, G.~Wan, Y.~Zhou, X.~Fu, P.~Yuan, and S.~Song, ``Deepvcp: An end-to-end
  deep neural network for point cloud registration,'' in {\em Proceedings of
  the IEEE International Conference on Computer Vision}, pp.~12--21, 2019.

\bibitem{zhou2018voxelnet}
Y.~Zhou and O.~Tuzel, ``Voxelnet: End-to-end learning for point cloud based 3d
  object detection,'' in {\em Proceedings of the IEEE Conference on Computer
  Vision and Pattern Recognition}, pp.~4490--4499, 2018.

\bibitem{behley2019semantickitti}
J.~Behley, M.~Garbade, A.~Milioto, J.~Quenzel, S.~Behnke, C.~Stachniss, and
  J.~Gall, ``Semantickitti: A dataset for semantic scene understanding of lidar
  sequences,'' in {\em Proceedings of the IEEE International Conference on
  Computer Vision}, pp.~9297--9307, 2019.

\bibitem{shan2018lego}
T.~Shan and B.~Englot, ``Lego-loam: Lightweight and ground-optimized lidar
  odometry and mapping on variable terrain,'' in {\em 2018 IEEE/RSJ
  International Conference on Intelligent Robots and Systems (IROS)},
  pp.~4758--4765, 2018.

\bibitem{9197261}
Y.~{Yue}, C.~{Zhao}, R.~{Li}, C.~{Yang}, J.~{Zhang}, M.~{Wen}, Y.~{Wang}, and
  D.~{Wang}, ``A hierarchical framework for collaborative probabilistic
  semantic mapping,'' in {\em 2020 IEEE International Conference on Robotics
  and Automation (ICRA)}, pp.~9659--9665, 2020.

\bibitem{petrovskaya2009model}
A.~Petrovskaya and S.~Thrun, ``Model based vehicle detection and tracking for
  autonomous urban driving,'' {\em Autonomous Robots}, vol.~26, no.~2-3,
  pp.~123--139, 2009.

\bibitem{wojke2012moving}
N.~Wojke and M.~H{\"a}selich, ``Moving vehicle detection and tracking in
  unstructured environments,'' in {\em 2012 IEEE International Conference on
  Robotics and Automation}, pp.~3082--3087, IEEE, 2012.

\bibitem{tanzmeister2014grid}
G.~Tanzmeister, J.~Thomas, D.~Wollherr, and M.~Buss, ``Grid-based mapping and
  tracking in dynamic environments using a uniform evidential environment
  representation,'' in {\em 2014 IEEE International Conference on Robotics and
  Automation (ICRA)}, pp.~6090--6095, 2014.

\bibitem{nuss2018random}
D.~Nuss, S.~Reuter, M.~Thom, T.~Yuan, G.~Krehl, M.~Maile, A.~Gern, and
  K.~Dietmayer, ``A random finite set approach for dynamic occupancy grid maps
  with real-time application,'' {\em The International Journal of Robotics
  Research}, vol.~37, no.~8, pp.~841--866, 2018.

\bibitem{wu2020motionnet}
P.~Wu, S.~Chen, and D.~N. Metaxas, ``Motionnet: Joint perception and motion
  prediction for autonomous driving based on bird's eye view maps,'' in {\em
  Proceedings of the IEEE/CVF Conference on Computer Vision and Pattern
  Recognition}, pp.~11385--11395, 2020.

\bibitem{darms2009obstacle}
M.~S. Darms, P.~E. Rybski, C.~Baker, and C.~Urmson, ``Obstacle detection and
  tracking for the urban challenge,'' {\em IEEE Transactions on intelligent
  transportation systems}, vol.~10, no.~3, pp.~475--485, 2009.

\bibitem{cho2014multi}
H.~Cho, Y.-W. Seo, B.~V. Kumar, and R.~R. Rajkumar, ``A multi-sensor fusion
  system for moving object detection and tracking in urban driving
  environments,'' in {\em 2014 IEEE International Conference on Robotics and
  Automation (ICRA)}, pp.~1836--1843, 2014.

\bibitem{mertz2013moving}
C.~Mertz, L.~E. Navarro-Serment, R.~MacLachlan, P.~Rybski, A.~Steinfeld,
  A.~Suppe, C.~Urmson, N.~Vandapel, M.~Hebert, C.~Thorpe, {\em et~al.},
  ``Moving object detection with laser scanners,'' {\em Journal of Field
  Robotics}, vol.~30, no.~1, pp.~17--43, 2013.

\bibitem{apollo}
Baidu.
\newblock \url{https://github.com/ApolloAuto/apollo}.

\bibitem{held2016robust}
D.~Held, J.~Levinson, S.~Thrun, and S.~Savarese, ``Robust real-time tracking
  combining 3d shape, color, and motion,'' {\em The International Journal of
  Robotics Research}, vol.~35, no.~1-3, pp.~30--49, 2016.

\bibitem{wang2019precise}
D.~Wang, J.~Xue, W.~Zhan, Y.~Jin, N.~Zheng, and M.~Tomizuka, ``Precise
  correntropy-based 3d object modelling with geometrical traffic prior,'' in
  {\em 2019 IEEE/RSJ International Conference on Intelligent Robots and Systems
  (IROS)}, pp.~2608--2613, 2019.

\bibitem{yang2019observability}
Y.~Yang and G.~Huang, ``Observability analysis of aided ins with heterogeneous
  features of points, lines, and planes,'' {\em IEEE Transactions on Robotics},
  vol.~35, no.~6, pp.~1399--1418, 2019.

\bibitem{geneva2018lips}
P.~Geneva, K.~Eckenhoff, Y.~Yang, and G.~Huang, ``Lips: Lidar-inertial 3d plane
  slam,'' in {\em 2018 IEEE/RSJ International Conference on Intelligent Robots
  and Systems (IROS)}, pp.~123--130, 2018.

\bibitem{van2000computational}
M.~Van~Kreveld, O.~Schwarzkopf, M.~de~Berg, and M.~Overmars, {\em Computational
  geometry algorithms and applications, 3rd Edition}.
\newblock Springer, 2008.

\bibitem{bogoslavskyi2016fast}
I.~Bogoslavskyi and C.~Stachniss, ``Fast range image-based segmentation of
  sparse 3d laser scans for online operation,'' in {\em 2016 IEEE/RSJ
  International Conference on Intelligent Robots and Systems (IROS)},
  pp.~163--169, IEEE, 2016.

\bibitem{fisher1995statistical}
N.~I. Fisher, {\em Statistical analysis of circular data}.
\newblock cambridge university press, 1995.

\bibitem{berens2009circstat}
P.~Berens {\em et~al.}, ``Circstat: a matlab toolbox for circular statistics,''
  {\em J Stat Softw}, vol.~31, no.~10, pp.~1--21, 2009.

\bibitem{forney1973viterbi}
G.~D. Forney, ``The viterbi algorithm,'' {\em Proceedings of the IEEE},
  vol.~61, no.~3, pp.~268--278, 1973.

\bibitem{hash-map}
MIT.
\newblock \url{https://github.com/Tessil/robin-map}.

\end{thebibliography}

\end{document}